\documentclass[10pt,letterpaper]{article}
\usepackage{graphicx}
\usepackage{color}

\usepackage[top=0.85in,left=2.75in,footskip=0.75in]{geometry}

\usepackage{amsmath,amssymb}

\usepackage{changepage}

\usepackage{textcomp,marvosym}

\usepackage{cite}

\usepackage{nameref,hyperref}

\usepackage[nopatch=eqnum]{microtype}
\DisableLigatures[f]{encoding = *, family = * }

\usepackage[table]{xcolor}

\usepackage{array}

\newcolumntype{+}{!{\vrule width 2pt}}

\newlength\savedwidth

\raggedright
\usepackage[aboveskip=1pt,labelfont=bf,labelsep=period,justification=raggedright,singlelinecheck=off]{caption}

\makeatletter
\renewcommand{\@biblabel}[1]{\quad#1.}
\makeatother

\usepackage{lastpage,fancyhdr,graphicx}
\usepackage{epstopdf}
\fancyheadoffset[L]{2.25in}
\fancyfootoffset[L]{2.25in}
\begin{document}
\vspace*{0.2in}

\begin{flushleft}
{\Large
\textbf\newline{Cyclic image generation using chaotic dynamics} % Please use "sentence case" for title and headings (capitalize only the first word in a title (or heading), the first word in a subtitle (or subheading), and any proper nouns).
}
\newline
\\
Takaya Tanaka\textsuperscript{1},
Yutaka Yamaguti\textsuperscript{2*},
\\
\bigskip
\textbf{1} Graduate school of Engineering, Fukuoka Institute of Technology, Fukuoka, Fukuoka, Japan
\\
\textbf{2} Faculty of Information Engineering, Fukuoka Institute of Technology, Fukuoka, Fukuoka, Japan
\\
\bigskip

% Use the asterisk to denote corresponding authorship and provide email address in note below.
* y-yamaguchi@fit.ac.jp

\end{flushleft}
\section*{Abstract}

Successive image generation using cyclic transformations is demonstrated by extending the 
CycleGAN model to transform images among three different categories. 
Repeated application of the trained generators produces sequences of images that 
transition among the different categories. The generated image sequences occupy a more 
limited region of the image space compared with the original training dataset. 
Quantitative evaluation using precision and recall metrics indicates that the generated 
images have high quality but reduced diversity relative to the training dataset. 
Such successive generation processes are characterized as chaotic dynamics in terms of 
dynamical system theory. Positive Lyapunov exponents estimated from the generated trajectories 
confirm the presence of chaotic dynamics, with the Lyapunov dimension of the attractor found to 
be comparable to the intrinsic dimension of the training data manifold. The results suggest that 
chaotic dynamics in the image space defined by the deep generative model contribute to the 
diversity of the generated images, constituting a novel approach for multi-class image generation. 
This model can be interpreted as an extension of classical associative memory to perform 
hetero-association among image categories.

\section*{Author summary}
We have developed a new approach to generate sequences of related images by extending a type of deep learning model called CycleGAN. Our model learns to transform images among three different categories in a cyclic manner. For example, it can turn an image of a T-shirt into a sneaker, then a bag, and back to a T-shirt. Our model generates a series of smoothly changing images by repeatedly applying these transformations. Interestingly, we found that the sequences of generated images exhibit chaotic dynamics. This means that even tiny changes to the starting image can lead to very different sequences. We characterized these chaotic dynamics quantitatively using metrics from chaos theory. Our results demonstrate a novel way to generate diverse images related to multiple categories. The generated images are of high quality but tend to be less diverse than the original training data. Our model may serve as an interesting model of associative memory from a theoretical neuroscience perspective.
%\linenumbers

\section*{Introduction}
\label{sec:introduction}
As deep learning has advanced, there has been extensive research into models that can generate realistic images~\cite{goodfellow2014,kingma2013auto,ho2020denoising}, and one such model is the generative adversarial network (GAN)~\cite{goodfellow2014}. 
A GAN consists of two networks---the generator and the discriminator---that are trained adversarially 
to generate new images that are similar to those in the training dataset. 
This has led to the development of various extended models, 
one of which is CycleGAN~\cite{zhu2017unpaired}, 
which is an advanced GAN that performs image translation by learning the relationship 
between two different image categories.

Elsewhere, associative memory models have been studied as models of biological memory in which data related to a given input pattern are stored and retrieved~\cite{anderson1972simple,nakano1972associatron,kohonen1972correlation,hopfield1982neural}.
Such memory can be realized by Hebbian-type synaptic learning.
As an extension of associative memory models,  
dynamic associative memory models that retrieve stored memory patterns by using chaotic dynamics 
were proposed from the late 1980s to the 1990s~\cite{tsuda1987, nara1992chaotic, adachi1997associative}.
It has been shown that these associative memory models based on chaotic dynamics can autonomously generate sequential patterns that resemble memory patterns. 
From the perspective of dynamical systems, each memory pattern is a pseudo-attractor; that is, the state remains near the pattern for a while and then transitions chaotically to another pattern. 
Such dynamics is called chaotic itinerancy~\cite{tsuda1992,kaneko2001book,tsuda2015}.
However, there has been limited research into using modern deep neural networks to generate successive similar images autonomously.

In machine learning, chaotic dynamical systems have been utilized in various ways. For example, chaotic time series are often used as benchmarks for evaluating the performance of time-series prediction models~\cite{gilpin2021chaos}. Moreover, Tanaka and Yamaguti~\cite{tanakayamaguti2023} showed that a GAN can be trained to generate chaotic time series, and they evaluated the properties of the generated time series from the perspective of deterministic chaos. In the context of reservoir computing using recurrent neural networks, maintaining a weak chaotic state in the absence of input has been shown to enhance the learning of complex behaviors~\cite{sussillo2009,laje2013robust}. 

However, despite these advancements, there has been limited research into constructing associative memory models that leverage chaotic dynamics within deep learning frameworks. Our study aims to fill this gap by proposing a novel approach that utilizes the power of chaos to explore the high-dimensional space of complex datasets. We construct a model that combines the image-transformation ability of deep learning with the autonomous and rich pattern-generation ability of chaotic dynamics to successively and associatively generate diverse images. Specifically, we develop a model that is an extension of CycleGAN~\cite{zhu2017unpaired}, and we generate images iteratively by feeding those produced by the model back into the same model.
While the conventional version of CycleGAN transforms images between two different image domains defined by two datasets, we construct a model that transforms images cyclically among three different domains.
This successive image generation can be treated as a dynamical system defined in the image space, and so its behavior can be analyzed using concepts from nonlinear dynamical systems, such as attractors and chaos. Specifically, we use Lyapunov exponents and Lyapunov dimensions to quantify the characteristics of this dynamical system, then we relate those characteristics to those of the image generation.
Our results show that this model can generate a wide range of images by utilizing chaotic dynamics. Moreover, the quality and diversity of the generated images are evaluated using precision and recall metrics~\cite{kynkaanniemi2019}. 

\section*{Background}
\label{sec:background}

In this section, we provide some background to this study. 
Since this research is an interdisciplinary study involving  machine learning and chaotic dynamical systems, we explain relevant basic concepts from both disciplines for a broad audience.

\subsection*{Generative adversarial network}
\label{sec:gan}
A GAN~\cite{goodfellow2014} is a generative deep learning model that consists of two networks, namely, a generator and a discriminator. A characteristic of a GAN is that its generator and discriminator are trained adversarially. The generator generates data from given latent variables. The discriminator takes in either training data or data generated by the generator and outputs the probability that the input data are from the training data.

The details of the GAN are as follows. Let $X = \{x_1, \ldots , x_M\}$ be a set of data (dataset) existing in the space of images $J = I^N$ with $N$ pixels, where $I = [-1, 1]$ is the interval from $-1$ to $1$. Here, $X$ is assumed to be a set of data independently sampled as $x \sim p_{\text{data}}(x)$ according to a probability distribution $p_{\text{data}}(x)$ on $J$. Let $z$ be a latent variable sampled from a latent space $J_z$ according to a known prior distribution $p_{z}$ on $J_z$. The generator $G$ is a mapping from $J_z \rightarrow J$ and the discriminator $D$ is a mapping from $J \rightarrow [0,1]$, and they have internal adjustable parameters $\theta_G$ and $\theta_D$, respectively. The objective function $V$ of the GAN is given by 
\begin{equation}
\min_{\theta_G} \max_{\theta_D} V(D,G) = \min_{\theta_G} \max_{\theta_D} \left( \mathbb E_{x \sim p_{\text{data}}(x)}[\log{D(x)}]+ \mathbb E_{z \sim p_z(z)}[\log{(1-D(G(z)))}] \right), \label{eq:Objective_function_of_GAN}
\end{equation}
where $\mathbb E$ represents the expected value. 

When implementing $G$ and $D$ using neural networks and performing minimax training represented by Eq~\ref{eq:Objective_function_of_GAN} , 
the training phases of $G$ and $D$ are executed alternately.
In each training iteration, the generator optimizes its internal parameters to improve its generation ability so that the discriminator will mistakenly recognize the generated data as being real. 
Conversely, the discriminator optimizes its internal parameters to distinguish between real and fake data. This step is repeated alternately.

\subsection*{CycleGAN}\label{section:CycleGAN}

As an extension of a GAN, CycleGAN~\cite{zhu2017unpaired} learns the relationship between two unpaired image datasets and then can translate images between them.
For example, it can transform a photograph of an actual landscape into a painting of that landscape in the style of a particular artist.
Let $J = I^N$ be the space of images with $N$ pixels, where $I = [-1, 1]$, and suppose that there are two different datasets $X = \{x_1, x_2, \ldots , x_{M_X}\}$ and $Y = \{y_1, y_2, \ldots , y_{M_Y}\}$ on $J$. It is assumed that each data point in $X$ and $Y$ is independently sampled from the probability distributions $P_X$ and $P_Y$ on $J$, respectively. CycleGAN has two generators, namely, $G: J \rightarrow J$ and $F: J \rightarrow J$, and two discriminators, namely, $D_X: J \rightarrow [0,1]$ and $D_Y: J \rightarrow [0,1]$. Generator $G$ learns the transformation from $X$ to $Y$, while generator $F$ learns the transformation from $Y$ to $X$. Discriminator $D_Y$ (resp.\ $D_X$) distinguishes whether the input data were generated by generator $G$ (resp.\ $F$) or came from dataset $Y$ (resp.\ $X$).

The adversarial loss used to train the generators and discriminators is the same as in a conventional GAN, except that the generators' input is a sample image. The loss for $G$ and $D_Y$ is expressed as
\begin{equation}
\begin{split}
\mathcal{L}_{\text{GAN}}(G,D_Y,X,Y) &= \mathbb{E}_{y \sim P_Y}[\log D_Y(y)] \\
&+ \mathbb{E}_{x\sim P_X}[\log (1-D_Y(G(x)))].
\label{eq:Adversarial_loss}
\end{split}
\end{equation}
However, using only the adversarial loss, the learning of $G$ and $F$ remains independent, and the mutual conversion between domains may not converge well. To address this issue, an additional loss term is introduced to encourage $G$ and $F$ to approach an inverse mapping relationship, i.e., ${F}({G}({x}))\approx {x}$.
Called the cycle-consistency loss, this term is defined as 
\begin{equation}
\begin{split}
\mathcal{L}_{\text{cyc}}(G,F)&=\mathbb{E}_{x \sim p_{\text{data}}(x)}[\|F(G(x))-x\|_1] \\
&+ \mathbb{E}_{y \sim p_{\text{data}}(y)}[\|G(F(y))-y\|_1].
\label{eq:Cycle-consistency_loss_function}
\end{split}
\end{equation}
The cycle-consistency loss ensures that the learned transformations $G$ and $F$ are consistent with each other, enabling the model to capture the underlying correspondence between the two domains.
Here, $\|$ $\|_1$ denotes the L1 norm, also known as the Manhattan distance. 
The total objective function is defined by combining the adversarial loss and the cycle-consistency loss:
\begin{equation}
  \begin{split}
    \mathcal{L}(G,F,D_X,D_Y)&=\mathcal{L}_{\text{GAN}}(G,D_Y,X,Y) \\
    &+\mathcal{L}_{\text{GAN}}(F,D_X,Y,X) \\
    &+\lambda\mathcal{L}_{\text{cyc}}(G,F),
    \label{eq:Objective_function_of_CycleGAN}
  \end{split}
\end{equation}
where $\lambda$ is a parameter to control the relative weights of the two losses.
Then, learning is performed to find the optimal solution that satisfies
\begin{equation}
  G^*, F^* = \arg \min_{G,F}\max_{D_X,D_Y} \mathcal{L}(G,F,D_X,D_Y),
  \label{eq:CycleGAN_Training}
\end{equation}
where $G^*$ and $F^*$ represent the optimal models with the optimal parameters.

\subsection*{Chaotic dynamics and Lyapunov exponents}

Chaos is the irregular and seemingly random behavior in deterministic dynamical systems~\cite{alligood2000chaos}. 
In the context of dynamical systems, a ``trajectory'' refers to the path that the system's state traces over time, starting from a given initial state.
A key characteristic of chaos is sensitive dependence on the initial state, meaning that even small differences in the initial state can lead to very different trajectories over time. 
This sensitivity is quantified by Lyapunov exponents, which measure the average rate of divergence between nearby trajectories.

Lyapunov exponents are important quantities in the characterization of chaos~\cite{alligood2000chaos}. In an $N$-dimensional dynamical system, there are generally $N$ Lyapunov exponents that depend on the initial conditions, and the set of these exponents is called the Lyapunov spectrum. 
In a dynamical system  defined by a map \(f: \mathbb{R}^{N} \to \mathbb{R}^{N}\) on an $N$-dimensional space as $x_{n+1} = f(x_n)$, $x_{n} \in \mathbb{R}^{N}$ $(n=0, 1, \ldots )$, the Lyapunov exponents are calculated as the average of the logarithm of the local expansion rate on the trajectory generated by the map \(f\). The Jacobian matrix \(M_n\) for $n$ iterations of $f$ is expressed as 
\[
M_n = \prod_{k=0}^{n-1} J_k,
\]
where $J_k$ is the Jacobian matrix $J_k=\frac{\partial }{\partial x} f(x_k)$ at \(x_k\). 
Under certain conditions, it can be shown that the following matrix exists~\cite{eckmann1985ergodic}:
\begin{equation}
\Lambda = \lim_{n \to \infty} { \left( M^{T}_n M_n \right)^{1 \over 2n} }.
\label{eq:Lyap}
\end{equation} 
The Lyapunov spectrum $\{\lambda_{i}\}$, $i=1, \ldots ,N$ is obtained as the set of logarithms of the eigenvalues of the matrix $\Lambda$.

In the present numerical calculations, the expansion rate was estimated by the commonly used method of Gram--Schmidt orthonormalization~\cite{shimada1979numerical}.
Although the present generator model has a deep structure, 
each layer of the model is composed of functions that are differentiable almost everywhere. Consequently, the entire map $G$ is also differentiable almost everywhere, and the Jacobian matrix can be obtained numerically.
The Jacobian matrices were calculated numerically by using the ``batch\_jacobian'' function 
in the TensorFlow~\cite{tensorflow2015} library.
 
\section*{Model and methods}\label{chapter:model}
\subsection*{Model} 

\begin{figure}[!h]
  \centering
  \includegraphics[scale=0.5]{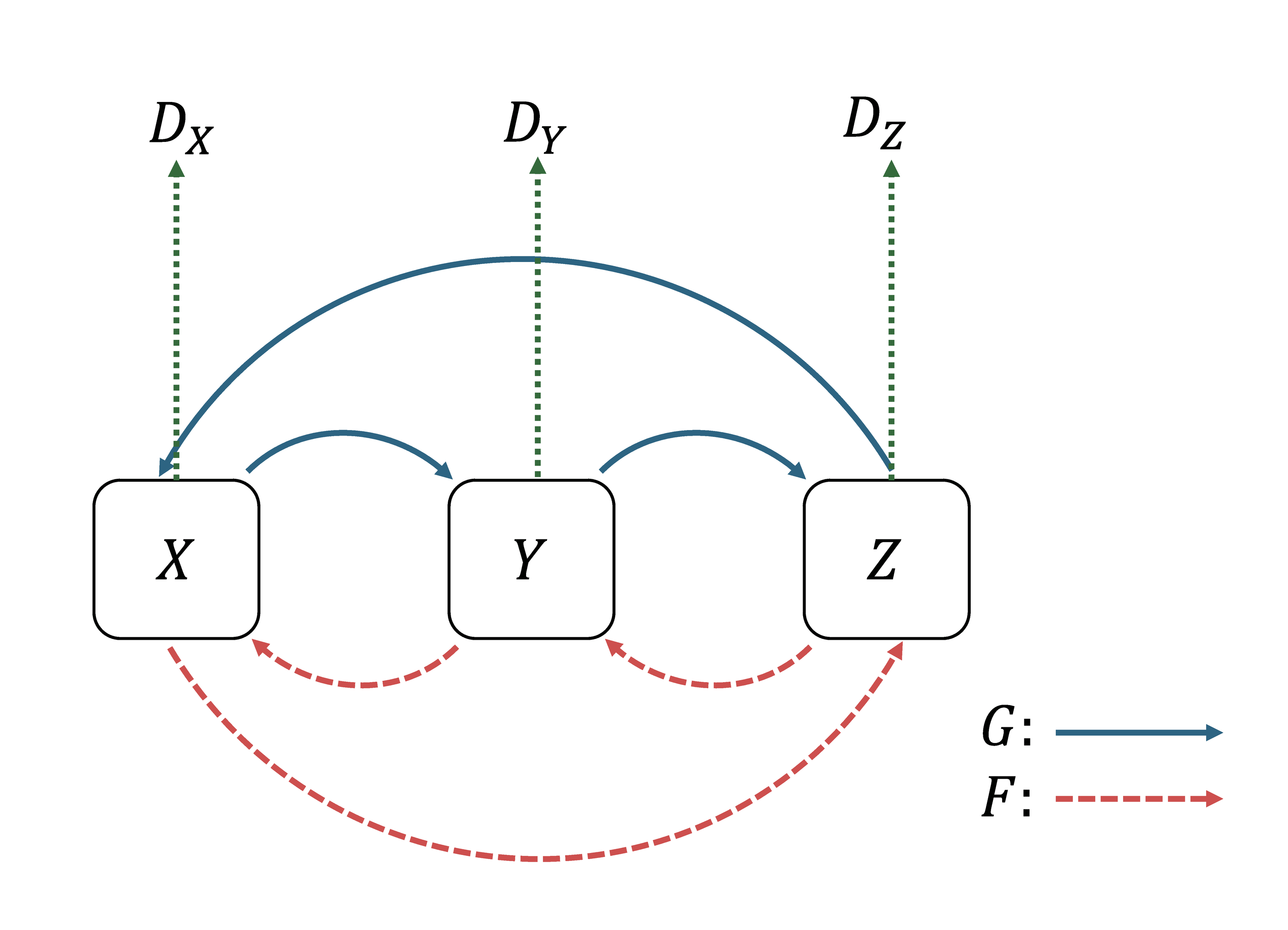}
  \caption{\bf Transformations learned by model.}
  The model consists of two generators, $G$ and $F$, and three discriminators, $D_X$, $D_Y$, and $D_Z$.
  \label{fig:model}
\end{figure}

We have developed a model that extends CycleGAN~\cite{zhu2017unpaired} to transform input images into different images cyclically, as shown in Fig~\ref{fig:model}, where $X$, $Y$, and $Z$ represent different image categories within the image space $J$.
This model has two generators ($G$ and $F$), and the transformations that they learn are represented by the solid ($G$) and dashed ($F$) arrows in Fig~\ref{fig:model}. 
Additionally, the model incorporates three discriminators: $D_X$, $D_Y$, and $D_Z$, each corresponding to one of the image categories.
Similar to CycleGAN,$G$ and $F$ are trained so that they are inverse maps of each other.

Below, we provide a brief overview of the network structures of the generator and discriminator used in this experiment, with more details available in \nameref{S1_Fig}  and \nameref{S2_Fig} in the supporting information. 
In the generator, the image is gradually downsized using convolution layers and then upsampled using transposed convolution layers to restore it to its original size. 
Between the downsampling and upsampling layers, a residual network (ResNet)~\cite{he2016identity} is inserted.
The discriminator extracts features from the input image by downsampling it with convolution layers. 
The feature map is then averaged using a global average pooling layer and input to the dense layer, and finally a single value is output as the decision result. 
Dropout layers~\cite{srivastava2014dropout} are incorporated into both the generator and discriminator to prevent overfitting.
Note that while this results in stochastic behavior during training, that is the only time; when all the results reported in this study were obtained, the dropouts were disabled and all neurons were activated, resulting in deterministic behavior. 

The construction, training, image generation, and evaluation of the model were performed 
using Python~3 codes with TensorFlow~2.7~\cite{tensorflow2015} library. 
The model's computations were performed on an NVIDIA GeForce RTX 4090 GPU  with 32-bit floating-point precision. The codes used for computations and analyses are available from a public GitHub repository at https://github.com/yymgch/cycle-chaos-gan.

\subsection*{Training datasets}
In this study, we performed training using the MNIST~\cite{MNIST} and Fashion-MNIST~\cite{xiao2017fashion} datasets. 

MNIST is a widely used dataset consisting of $7\times 10^4$ grayscale images of handwritten digits from 0 to 9. 
Of these images, $6\times 10^4$ are provided for training and $10^4$ are provided for testing. 
Each image has a resolution of $28 \times 28$ pixels, resulting in a total of 784 pixels per image.
For our experiments with MNIST, we used subsets of images belonging to specific categories: images of the digit~0 were assigned to dataset $X$, images of the digit~1 were assigned to dataset $Y$, and images of the digit~2 were assigned to dataset $Z$ (Fig~\ref{fig:model}).

Fashion-MNIST~\cite{xiao2017fashion} is another dataset that serves as a drop-in replacement for MNIST.
It consists of $7\times 10^4$ grayscale images from 10 categories of fashion products. The dataset maintains the same image size and split as MNIST, with $6\times 10^4$ training images and $10^4$ testing images, each having a resolution of $28 \times 28$ pixels. 
We used subsets of images belonging to specific fashion product categories: images of T-shirts/tops were assigned to dataset $X$, images of sneakers were assigned to dataset $Y$, and images of bags were assigned to dataset $Z$ (Fig~\ref{fig:model}).

\subsection*{Loss function}

The loss function for training our model is defined by extending the loss function of CycleGAN~\cite{zhu2017unpaired}. For the $X\rightarrow Y$ transformation by generator $G$, the loss function $\mathcal{L}_{\text{GAN}}(G,D_Y,X,Y)$ defined in Eq~\ref{eq:Adversarial_loss} for CycleGAN was employed. Similarly, loss functions were prepared for the transformations $Y\rightarrow Z$ and $Z\rightarrow Y$, as well as for the reverse transformation $Y\rightarrow X\rightarrow Z\rightarrow Y$ using $F$. The overall adversarial loss $\mathcal{L}_{\text{ADV}}$ was the sum of these:
\begin{equation}
  \begin{split}
    \mathcal{L}_{\text{ADV}}(G, F, D_X, D_Y, D_Z) &= \mathcal{L}_{\text{GAN}}(G,D_Y,X,Y) +\mathcal{L}_{\text{GAN}}(F,D_X,Y,X) \\
    &+ \mathcal{L}_{\text{GAN}}(G,D_Z,Y,Z) +\mathcal{L}_{\text{GAN}}(F,D_Y,Z,Y) \\
    &+ \mathcal{L}_{\text{GAN}}(G,D_X,Z,X) +\mathcal{L}_{\text{GAN}}(F,D_X,X,Z). \\
        \label{eq:model_adversarial_loss}
  \end{split}
\end{equation}

Furthermore, the cycle-consistency loss, which evaluates how close $G$ and $F$ are to being inverse mappings of each other, is formulated as 
\begin{equation}
  \begin{split}
    \mathcal{L}_{\text{CYC}}(G, F) &= \mathbb{E}_{x \sim P_X}[\| F(G(x)) - x \|_1]+ \mathbb{E}_{y \sim P_Y}[\| G(F(y)) - y \|_1] \\
    &+ \mathbb{E}_{y \sim P_Y}[\| F(G(y)) - y \|_1] + \mathbb{E}_{z \sim P_Z}[\| G(F(z)) - z \|_1] \\
    &+ \mathbb{E}_{z \sim P_Z}[\| F(G(z)) - z \|_1] + \mathbb{E}_{x \sim P_X}[\| G(F(x)) - x \|_1].
    \label{eq:model_cycle_consistency_loss}
  \end{split}
\end{equation}

The total objective function is a combination of the adversarial loss and the cycle-consistency loss:
\begin{equation}
\mathcal{L}(G, F, D_X, D_Y, D_Z) = \mathcal{L}_{\text{ADV}}(G, F, D_X, D_Y, D_Z) + \lambda \mathcal{L}_{\text{CYC}}(G, F),
\label{eq:model_total_loss}
\end{equation}
where $\lambda$ is a coefficient that controls the relative weight of the two losses, and it was set to $10$ in our experiments.
It is important to note that although the loss function does not explicitly include a term to promote the generation of diverse images, a GAN with the adversarial loss learns to match the distribution of the mapped data (e.g., the distribution of $G(x)$, $x\sim P_X$) with the distribution of the corresponding dataset (e.g., $P_Y$)~\cite{goodfellow2014}. Because of this characteristic, this model is expected to produce a wide variety of images that resemble the distribution of the target dataset.

\subsection*{UMAP}
\label{sec:umap}

We used uniform manifold approximation and projection (UMAP)~\cite{mcinnes2018umap} to perform dimensionality reduction on the image data and visually evaluate the extent of the generated data's coverage of the training data distribution. 
UMAP is a technique that can map high-dimensional data into a low-dimensional space while preserving the original data's local structure. 
For the UMAP parameters, we used the default values recommended by a previous study~\cite{mcinnes2018umap}, i.e., the number of nearest neighbors was set to $15$, the minimum distance
to $0.1$, and the distance metric to Euclidean distance.
% With these parameter settings, we observed clustered distributions for each dataset when mapping the training data onto a 2D plane.

\subsection*{Precision and recall}
\label{sec:precision_recall}

To quantitatively evaluate the quality and diversity of the generated images, we used the precision and recall metrics proposed by Kynk\"{a}\"{a}nniemi et~al.~\cite{kynkaanniemi2019} for evaluating generated data. Precision assesses the extent to which the generated images resemble the actual dataset images, while recall measures the extent to which the generated images cover a wide range of features of the real data. These metrics allow us to determine how well the generated image sequences capture the features of the real data and exhibit diversity. In this method, for both the real and generated datasets, we construct explicit and non-parametric representations of the manifolds in which the data lie, independent of the model or parameters. These manifolds are then used to estimate the precision and recall of the generated image set.

Let $X_r$ be the set of real data samples and $X_g$ be the set of generated data samples. The samples in each set are embedded into a high-dimensional feature space using the pre-trained VGG16 model~\cite{simonyan2015a}. Pre-trained on the ILSVRC2012 dataset~\cite{ILSVRC15}, this model allows us to extract high-level features from the images. We used the feature map obtained from the layer before the final layer of the VGG16 model as the feature vector.

Let $\phi_r$ and $\phi_g$ represent the feature vectors extracted from a real image and a generated image, respectively, and let $\Phi_r$ and $\Phi_g$ represent the sets of feature vectors corresponding to $X_r$ and $X_g$. We took an equal number of samples from each set, i.e., $|\Phi_r| = |\Phi_g|$.

For each set of feature vectors \(\Phi \in \{\Phi_r, \Phi_g\}\), we define the corresponding manifold in the feature space. Specifically, we perform the following steps. First, for each feature vector in the set, we consider a hypersphere with a radius equal to the distance to its \(k\)-th nearest neighbor. Next, we define the manifold of the dataset $\Phi$ as the union of these hyperspheres.

To determine whether a given sample \(\phi\) lies within this manifold, we define the following binary function:
\begin{equation}
f(\phi, \Phi) = \begin{cases}
1 & \text{if } \|\phi - \phi'\|_2 \leq \|\phi' - \text{NN}_k(\phi', \Phi)\|_2 \text{ for at least one } \phi' \in \Phi ,\\
0 & \text{otherwise.}
\end{cases}
\label{eq:binary_function}
\end{equation}
Here, $\text{NN}_k (\phi', \Phi)$ is a function that returns the \(k\)-th nearest-neighbor feature vector to $\phi'$ from the set $\Phi$. Intuitively, $f(\phi, \Phi_r)$ determines whether a given image $\phi$ looks real, and $f(\phi, \Phi_g)$ determines whether a given image can be reproduced by the generator.

Precision measures how many generated images lie within the manifold of real images and is defined as 
\begin{equation}
\text{precision}(\Phi_r, \Phi_g) = \frac{1}{|\Phi_g|} \sum_{\phi_g \in \Phi_g} f(\phi_g, \Phi_r).
\label{eq:Precision}
\end{equation}
On the other hand, recall measures how many real images lie within the manifold of generated images and is defined as
\begin{equation}
\text{recall}(\Phi_r, \Phi_g) = \frac{1}{|\Phi_r|} \sum_{\phi_r \in \Phi_r} f(\phi_r, \Phi_g).
\label{eq:Recall}
\end{equation}
By using these metrics, we can quantitatively evaluate the quality and diversity of the generated image set.

\section*{Results}
\subsection*{Generation of image sequences}
To construct a dynamical system that iteratively generates various images in a cyclic manner, 
we built a deep generative model that sequentially transforms images from one of three categories into the next category, as described in the section entitled ``Model and methods.'' 
We trained the model using the loss function defined in Eq~\ref{eq:model_total_loss}, and the generators $G$ and $F$ were trained for 1000 epochs on the MNIST or Fashion-MNIST dataset. 
In the following sections, we analyze the model from the perspective of dynamical systems and evaluate the image sequences generated using the trained model.

\begin{figure}[!h]
  \centering
  \includegraphics[scale=0.2]{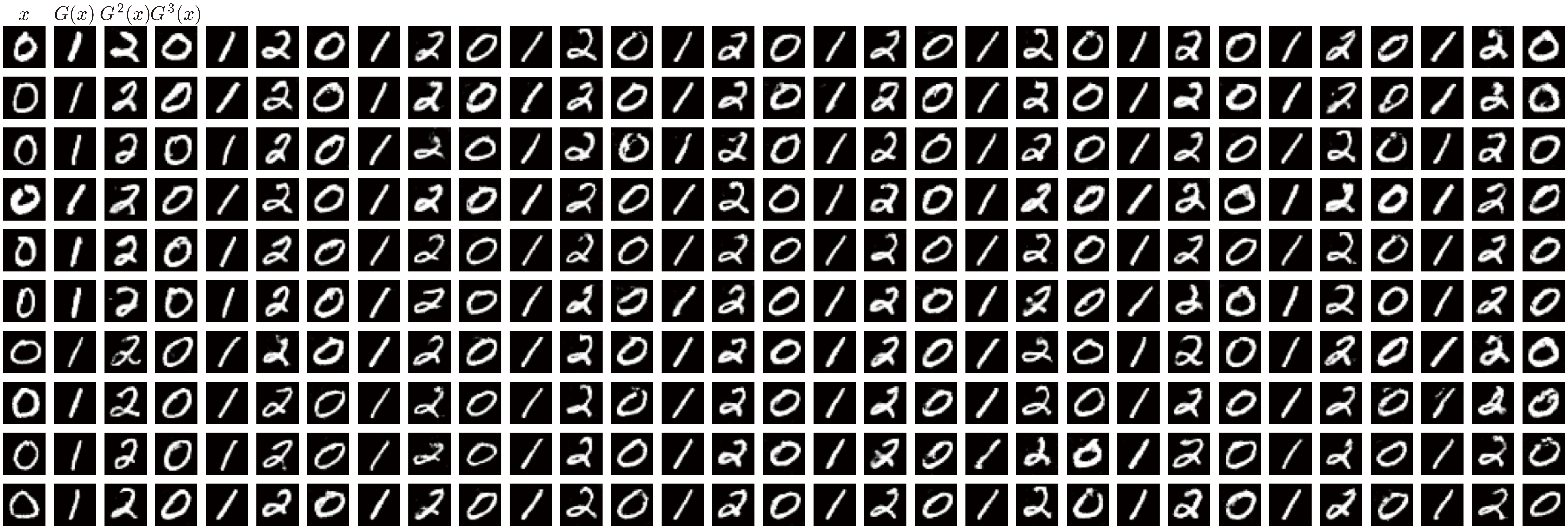}  
  \caption{{\bf Example of generated image sequences for MNIST.} 
  The leftmost image in each row is the initial image, and the subsequent images are generated by iteratively applying the generator $G$ to the previous image.
  }
  \label{fig:MNIST-samples}
\end{figure}

\begin{figure}[!h]
  \centering
  \includegraphics[scale=0.2]{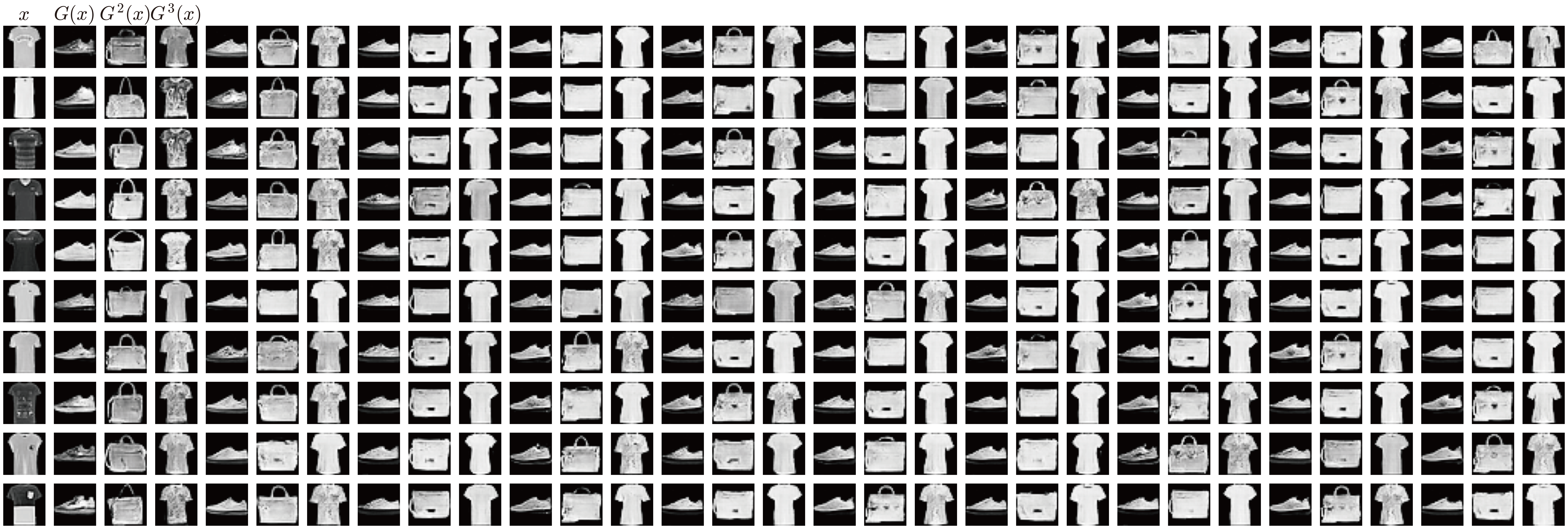} 
  \caption{{\bf Example of generated image sequences for Fashion-MNIST.} 
  The leftmost image in each row is the initial image, and the subsequent images are generated by iteratively applying the generator $G$ to the previous image.
  }
  \label{fig:Fashion-MNIST-samples}
\end{figure}

Fig~\ref{fig:MNIST-samples} shows the results of the iterative transformation from test image samples of the handwritten digit~0 as the initial values $x_0$, 
using the generator $G$ by applying $x_{n+1}=G(x_{n})$. 
Similarly, Fig~\ref{fig:Fashion-MNIST-samples} shows the results of the transformation of Fashion-MNIST images, starting from T-shirt images as the initial values. 
In both figures, multiple rows of images are shown. 
In each row, the leftmost image represents the initial value, and the subsequent images are generated by iteratively applying the generator to the previous image (left column) and displaying the result in the next column to the right. 
Each row shows a separate sequence of generated images starting from a different initial image. 
Both examples show that the images are transformed appropriately into the next digit or fashion product category.

\subsection*{Visualization of generated data distribution using UMAP}
\label{sec:UMAP}
\begin{figure}[h]
  \centering
  \includegraphics[scale=0.7]{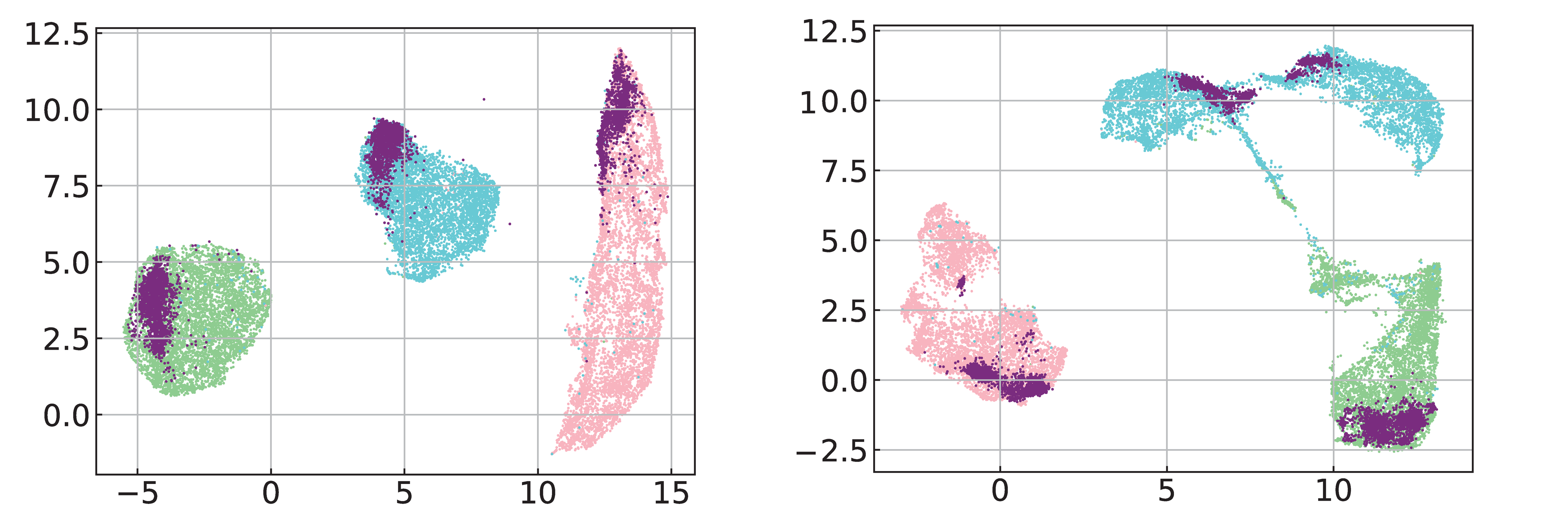}
  \caption{{\bf Visualization of distribution of training data and generated data by UMAP.} 
  Left: results for MNIST dataset. The green, pink, and cyan points represent 0, 1, and 2 image data in the training dataset, respectively, and the purple points represent the generated images. 
  Right: results for Fashion-MNIST dataset. The green, pink, and cyan points represent T-shirt/top, sneaker, and bag image data in the training dataset, respectively, and the purple points represent the generated images.}
  \label{fig:UMAP-traj-MNIST}
\end{figure}

To visually assess the distribution of the image sequences generated by the iteration of $G$, 
we visualize the distribution using UMAP~\cite{mcinnes2018umap}. 
We iterated the transformation 5000 times from an initial image and visualized the distribution of 
the 5000 generated data points on a 2D plane (Fig~\ref{fig:UMAP-traj-MNIST} left and right). 
To embed, we first trained the UMAP model to embed the training data distribution into a 2D latent space, then we embedded the generated data into the same latent space using the learned model. 
In the figures, the training data distributions of datasets $X$, $Y$, and $Z$ (digits~0, 1, and 2 for MNIST and T-shirt/top, sneaker, and bag for Fashion-MNIST) are represented by cyan, pink, and green points, respectively, while the generated data are represented by purple points.
  Note that the quantitative analysis in this paper was not performed based on the data in the UMAP-transformed space. The analysis of Lyapunov exponents was performed in the original image space, and the precision/recall metrics were calculated using the feature space obtained by the VGG16 model.
  We use the UMAP-transformed space only for visualization.

There are three clusters in the distributions of the training data, corresponding to the $X$, $Y$, and $Z$ datasets.
The generated data points lie within the regions where the training data exist, which suggests that the generated images closely resemble the original images, indicating high quality. 
However, the generated data distribution does not fully cover the entire distribution of the training data, suggesting that only a portion of the original data was reproduced in the generation sequence. 
These conjectures about quality and diversity are analyzed quantitatively in a later subsection. 
\nameref{S3_Fig} in the supporting information shows 64 different image sequences from different initial values. It appears that trajectories from different initial values converge to very similar distributions. From the perspective of dynamical systems, this suggests that these trajectories converge to the same attractors.

  To evaluate whether the UMAP transformation effectively captures the distribution of data and the characteristics of the trajectories in the original space, we performed two analyses.
  First, we applied k-means clustering to the UMAP-transformed training data to confirm that UMAP projected the three categories into two-dimensional space while preserving their distinctions. The Adjusted Rand Score~\cite{hubert1985comparing,steinley2004properties}, a measure of clustering accuracy, was approximately $0.976$. This high score indicates that UMAP successfully projected the clusters of the original categories into two dimensions while maintaining their differences.
  Next, we employed the Trustworthiness and Continuity metrics~\cite{venna2001} to verify that the trajectories in the UMAP space effectively capture the high-dimensional dynamics. These metrics quantify the extent to which points in close proximity in high dimensions remain close in low dimensions, and vice versa. The metrics range from 0 to 1, with values closer to 1 indicating better preservation of the data structure. For the trajectories of our dynamical system, the Trustworthiness and Continuity metrics were approximately $0.942$ and $0.978$, respectively. These high values demonstrate that the UMAP representation effectively preserves the characteristics of the trajectories from the high-dimensional space.

The results in Fig~\ref{fig:UMAP-traj-MNIST} suggest that trajectories starting from an initial condition converge to limited regions within the image space. From the perspective of dynamical systems, this can be interpreted as convergence to attractors. Furthermore, \nameref{S3_Fig} indicates that trajectories from different initial conditions also converge to the same attractors. To gain a more detailed understanding of the convergence process to these attractors, we observe bundles of trajectories starting from a large number of initial conditions. Specifically, we track the transformation of the set of states at each time step and visualize the process of convergence to the attractors. We analyze whether trajectories from different initial conditions are drawn to similar regions and how the region of images that the generator can produce shrinks over time.

\begin{figure}[!h]
  \includegraphics[scale=0.6]{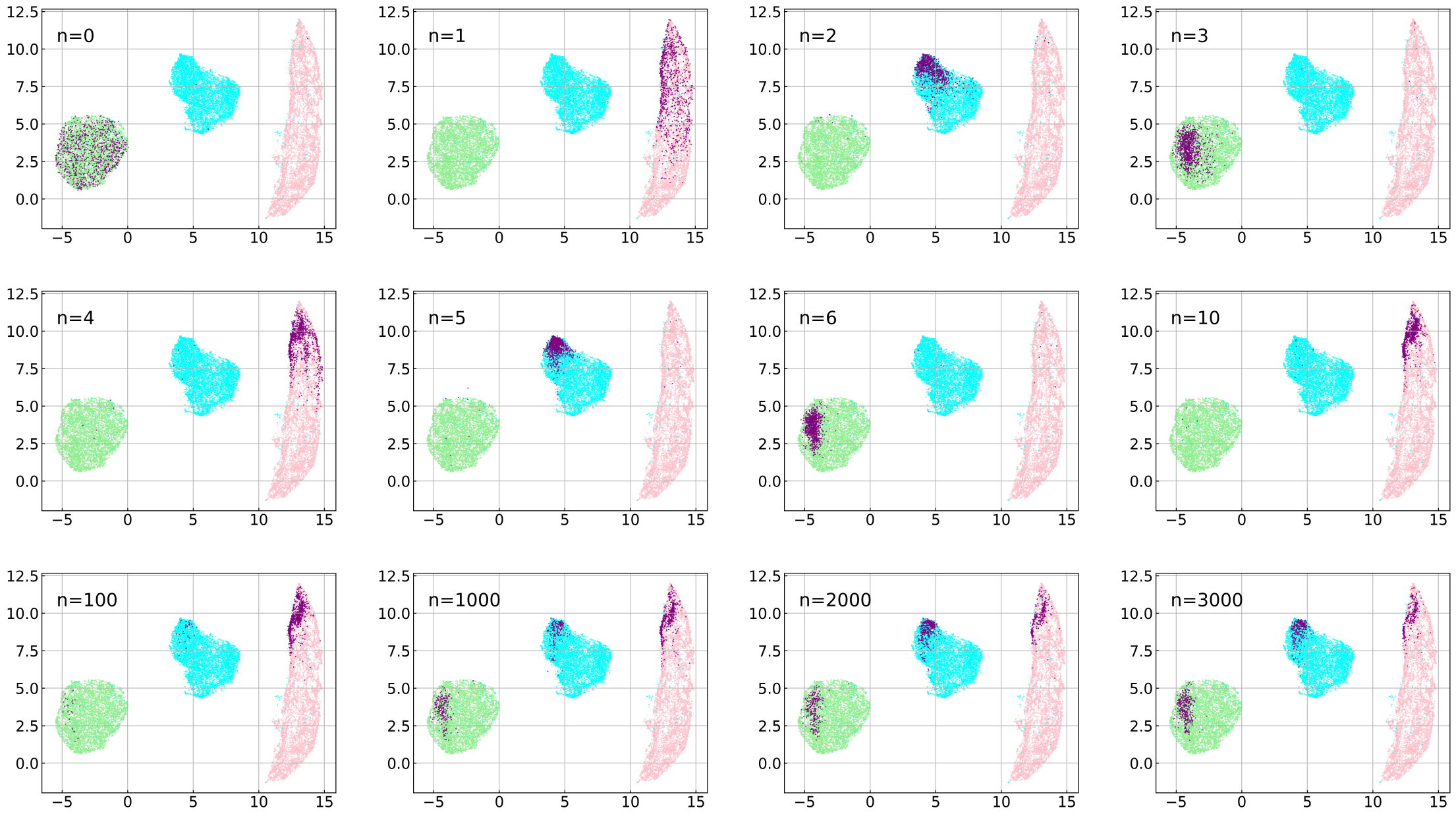}
  \caption{{\bf UMAP visualization of transitions of a set of states starting from a category $X$ image for MNIST dataset.}
  The step $n$ indicates the number of transformations applied to the initial image. Green, pink, and cyan points represent image data for digits 0, 1, and 2 in the training dataset, respectively, and purple points represent the generated images. The area where the transitioned points are present shrinks over time relative to the area where the training data exist.
  }
\label{fig:umap_many}
\end{figure}

We prepared a large number of initial values $\{x^{(1)}_0, \ldots , x^{(j)}_{0}, \ldots , x^{(980)}_{0}\}$ from test images of category $X$ and performed $3000$ transformations as $x^{(j)}_{n+1}=G(x^{(j)}_{n})$, $(j=1, \ldots , 980)$ using the trained generator $G$. For each time step $n$, we embedded the set of states $\{x^{(1)}_n, \ldots , x^{(980)}_n\}$ on a 2D plane using UMAP as shown in Fig~\ref{fig:umap_many}. Similar to Fig~\ref{fig:UMAP-traj-MNIST}, the points representing the training data for each dataset are shown in light colors, while the points representing the generated data are shown in purple. The distribution at $n=0$ represents the distribution of the test images, which almost entirely covers the spread of the training image distribution. In the first few transformations, the region occupied by the generated images shrinks and does not fully cover the training image distribution, although it covers a wider range compared with later time steps, indicating the generation of relatively diverse images. However, as the transformations are repeated, the covered region shrinks further, and the trajectories concentrate on a limited area. After a certain number of iterations, the shape of the region remains relatively stable, and the set of generated images still occupies a finite area even after a long time. This suggests that the attractors are neither fixed points nor periodic points but rather attractors with some spatial extent. To quantitatively support these observations, we perform an analysis using the precision and recall metrics in the next subsection.

% Place figure captions after the first paragraph in which they are cited.

\subsection*{Quantitative evaluation of quality and diversity}
\label{sec:result_precision_recall}
To quantitatively evaluate the quality and diversity of the generated image sequences, we calculated the precision and recall~\cite{kynkaanniemi2019} (P/R) of the generated images from the MNIST-trained model using the method described in the subsection entitled ``Precision and recall.'' 
Here, quality refers to how similar the generated images are to the real dataset images, whereas diversity refers to how well the generated images cover a wide range of variations in the real dataset. 
The important parameters in this method are the number of data samples and the number of nearest neighbors. 
In this method, an equal number of real and generated data samples should be prepared. 
In our case, the real data consisted of all the test dataset images from categories $X$, $Y$, and $Z$, amounting to $3178$ samples.
For the generated images, transformations were repeatedly applied starting from a test image $x_0$ selected from category $X$ as an initial value, resulting in a trajectory of 3178 points $\{x_1, \ldots , x_{3178}\}$, and these samples included in this trajectory were used for evaluation. 
Previous research~\cite{kynkaanniemi2019} proposed using $2\times 10^4$ samples each for real and generated data as a standard procedure. 
However, in this case, the number of test images does not reach this number, so direct comparison with other literature results is impossible. 
To ensure the validity of the evaluation, $3178$ samples from the same categories were extracted from the training dataset, and the P/R between the training and test datasets was calculated as a reference value. 
In the literature~\cite{kynkaanniemi2019}, the number of nearest neighbors $k$ is typically set to 3. However, because the number of data samples used in this study was different from that in the literature, it is difficult to draw conclusions from results with only a single value of $k$. Therefore, we varied $k$ from 1 to 10 and observed the resulting changes in P/R values.

\begin{figure}[!h]
  \centering
  \includegraphics[scale=0.7]{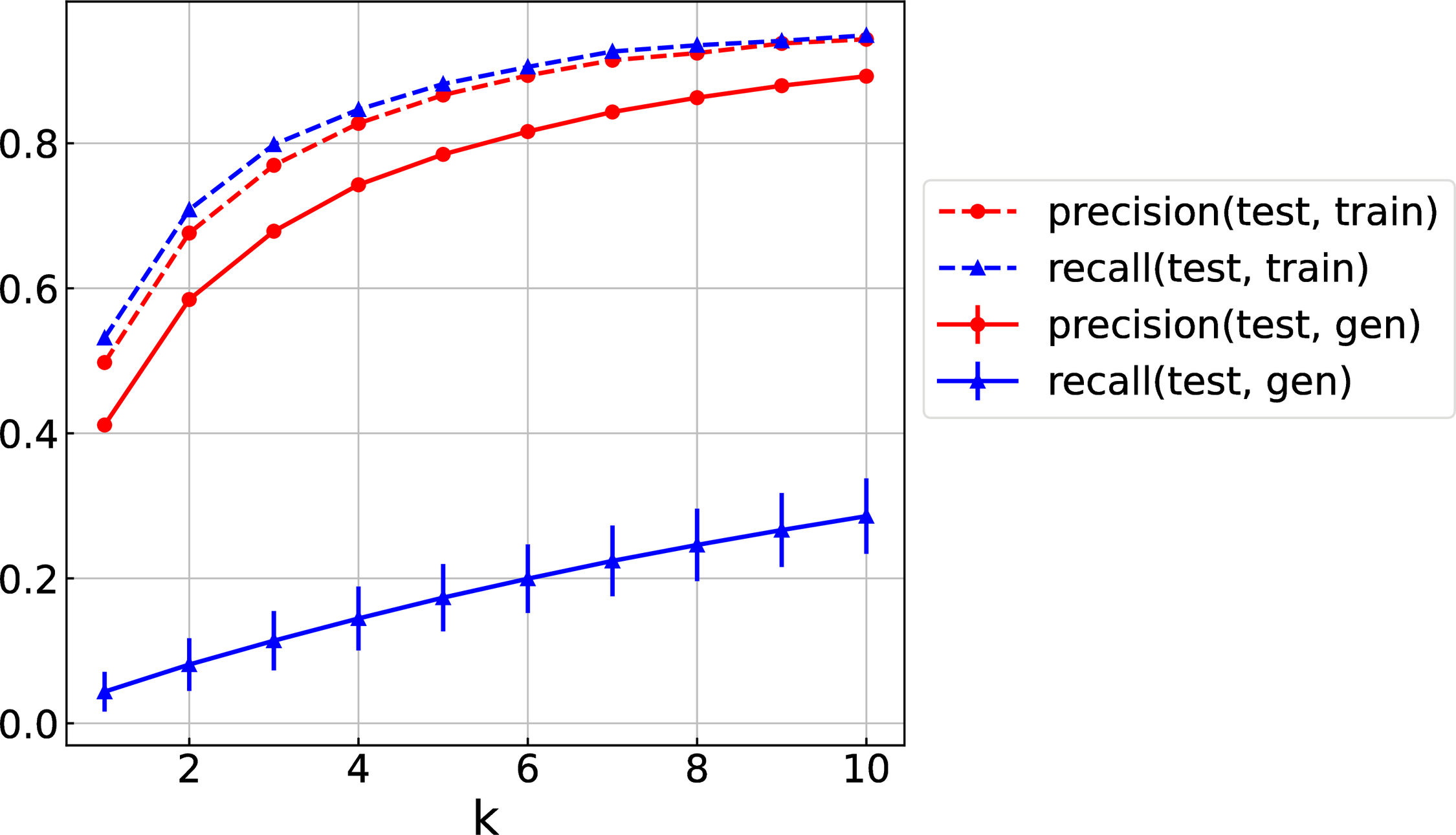}  
  \caption{{\bf Precision and recall of the generated image sequences.}
  The mean and standard deviation of the precision and recall values between 
  the test and generated data are shown by the solid red and blue lines, respectively, as functions of the number of nearest neighbors $k$.
  The P/R values between the training and test data are shown by the dashed lines as references.
}
\label{fig:precision_recall} 
\end{figure}

The P/R calculation results for each $k$ for the set of images included in the generated trajectory are shown in Fig~\ref{fig:precision_recall}. 
The P/R values between the test and generated data were calculated for 100 different trajectories, and their mean values and standard deviations are shown in the figure. 
These two values monotonically increase with $k$ because of the nature of the algorithm. 
The P/R values between the training and test data both rise to around 0.9 at around $k=6$, indicating high similarity between the training and test datasets. 
In contrast, while the precision between the generated and test data was slightly lower than that between the training and test data, the recall is considerably lower than that between the training and test data regardless of $k$. 
These results indicate quantitatively that while many of the generated images are of high quality and sufficiently close to the real images, the diversity of the generated images is lower than that of the real image dataset, and there are portions of the real images that are not reproduced in the generated image sequences.

\begin{figure}[!h]
  \centering
  \includegraphics[scale=0.45]{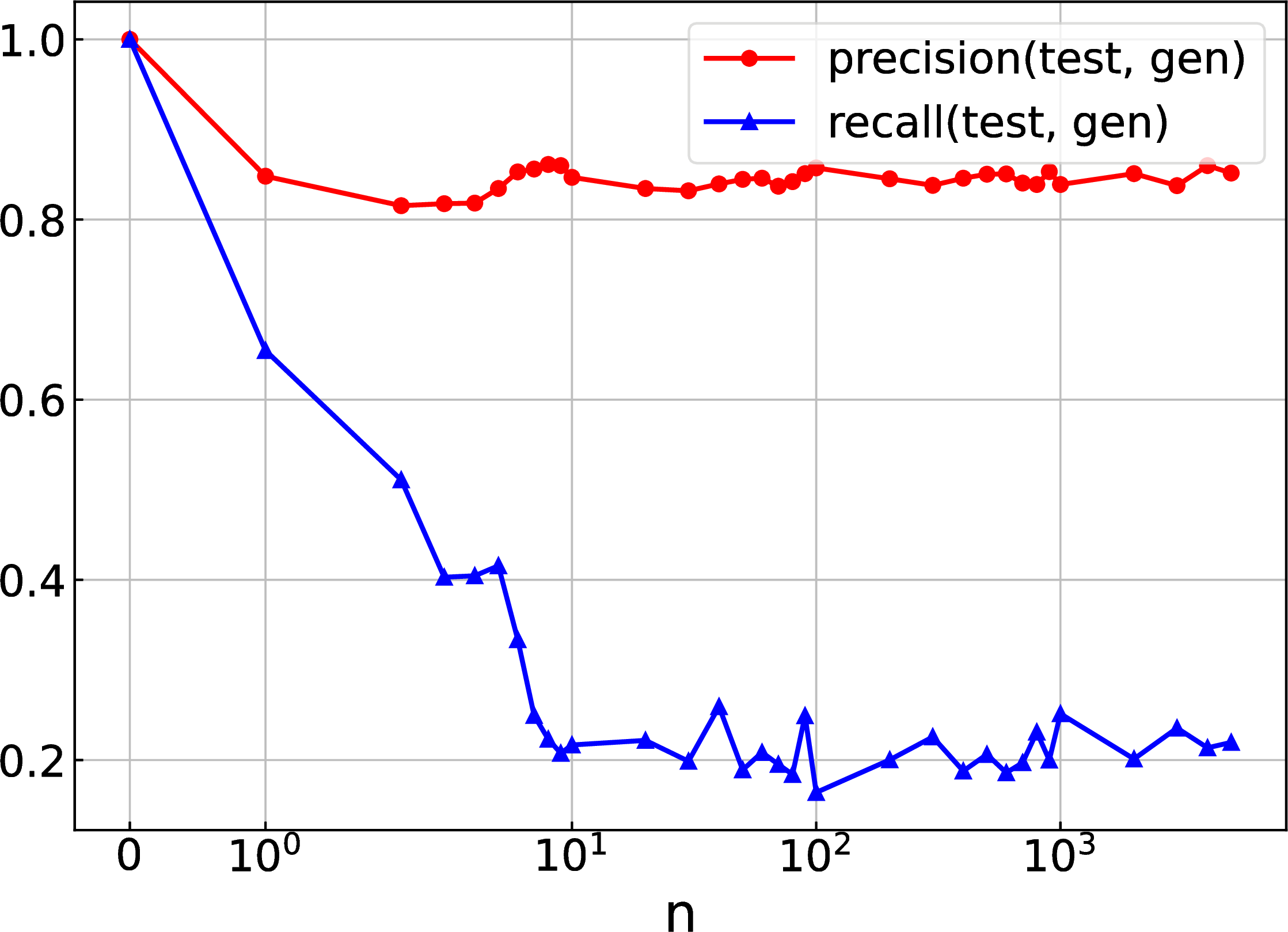}  
  \caption{{\bf Precision and recall as functions of time step.}
  Precision and recall of set of generated images as functions of the number of transformations applied to the initial image. The number of nearest neighbors was set to $k=7$.
  }
\label{fig:precision_recall_step}
\end{figure}

Next, we quantitatively evaluate how the images converge to a limited region as transformations are repeatedly applied, as shown in Fig~\ref{fig:umap_many}. 
Using the set of images obtained by mapping the entire set of test images for $n$ steps, we calculated the P/R for each step (Fig~\ref{fig:precision_recall_step}). 
Referring to the results in Fig~\ref{fig:precision_recall}, the number of nearest neighbors was set to $k=7$, where the P/R between the training and test data is sufficiently high.
The generated dataset at $n=0$ is the test dataset itself, and thus both the precision and recall are 1. 
For $n \geq 1$, the precision slightly decreases but remains high at above 0.8 for all steps. On the other hand, the recall decreases with $n$ to less than $0.3$ at $n=10$, and then fluctuates around that value. 
These results indicate that the decrease in diversity occurs gradually in the early stages of the iterations and then maintains a certain level of diversity for an extended period.

  To investigate the dependence of the number of  generated sequences on the evaluation of P/R values, we performed an additional analysis using a large number of artificially generated initial images and their trajectories (\nameref{S4_Fig}). While the P/R values changed quantitatively due to the increased number of data points, the overall trends remained consistent. Specifically, we observed persistent high Precision/low Recall patterns and an initial decrease in Recall over time.

\subsection*{Chaotic dynamics}
\label{sec:chaotic_dynamics}

\subsubsection*{Lyapunov spectrum}

As observed above, the trajectories of the generated images do not converge to fixed points or periodic points but instead generate various images. 
This diverse image generation is not possible with simple dynamics such as fixed-point or periodic attractors, suggesting the presence of chaotic dynamics.  To determine whether chaotic dynamics are present, the Lyapunov exponent, which quantifies the chaotic characteristics of the trajectory of the dynamical system produced by the generator $G$, was estimated numerically.
In the numerical calculations, we used the method described in the subsection entitled ``Chaotic dynamics and Lyapunov exponents,'' which involves calculating the Jacobian matrix and utilizing Gram--Schmidt orthonormalization to estimate all the Lyapunov exponents (called the Lyapunov spectrum). 
The spectrum was estimated by calculating the exponents from $980$ trajectories of length $2000$ and taking the sample mean to obtain the full set of Lyapunov exponents. 
To remove the transient period, we first applied the mapping $2000$ times to the initial images, and then used the subsequent 2000-step time series $\{x_{2000}, \ldots , x_{3999} \}$ to estimate the Lyapunov exponents.

\nameref{S5_Fig} shows the histograms of the first five Lyapunov exponents calculated for each of the 980 trajectories. The histograms exhibit unimodal Gaussian-like distributions, suggesting that these trajectories converge to the same attractor without multi-stability. Therefore, it is reasonable to average these values to estimate the Lyapunov exponents.

\begin{figure}[!h]
  \centering
  \includegraphics[scale=0.6]{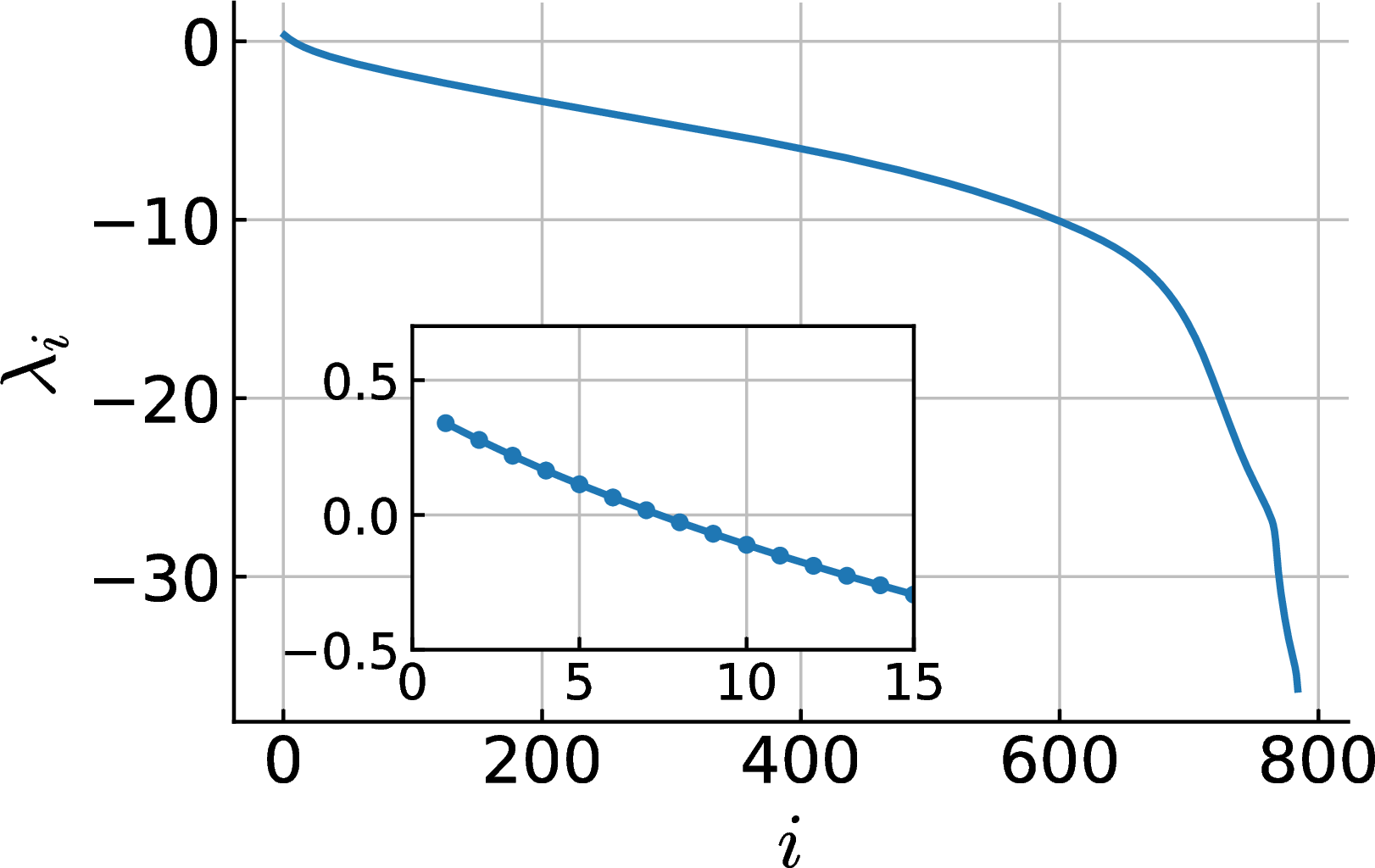}
  \caption{{\bf Lyapunov exponents of dynamics defined by generator $G$.}
  Shown here is the full Lyapunov spectrum, with the inset showing an enlarged view of the first 15 Lyapunov exponents.
  }
\label{fig:lyapunov}
\end{figure}

Fig~\ref{fig:lyapunov}  shows all the spectrum values, with the inset showing an enlarged view of the first 15 Lyapunov exponents. The first seven Lyapunov exponents are definitely positive, and the largest Lyapunov exponent is estimated to be about 0.340. The presence of these positive Lyapunov exponents indicates that the generated trajectories exhibit sensitive dependence on initial conditions and are chaotic, suggesting that the chaotic attractor generates various images.
  Most Lyapunov exponents are negative except for the seven largest. 
  The phenomenon of most exponents becoming negative is commonly observed in large-scale, high-dimensional dynamical systems where the dimension of the attractor is considerably smaller than the dimension of the system's phase space~\cite{kaneko2001book,engelken2023lyapunov,kobayashi2024lyapunov}.

  Further, \nameref{S6_Fig} demonstrates the convergence of estimated Lyapunov exponents over time, obtained by calculating trajectories from 10 initial values for an extended period ($2\times 10^6$ steps). 
  All ten trajectories converge to almost the same value, which matches the average value shown in Fig \ref{fig:lyapunov}.
  This result provides an additional evidence for the stability of our Lyapunov exponent calculations in this model.

\subsubsection*{Direct observation of trajectory instability}
It is well known that numerical computations of certain chaotic dynamical systems can be unstable~\cite{kaneko2001book}. Because our deep model requires complex computations with many parameters and these computations are performed using a GPU with finite precision, it is desirable to check the robustness of the numerical results for the Lyapunov exponents estimated in the previous subsection. To do so, we estimate the Lyapunov exponents using a different approach and then check the consistency of the results. For this purpose, we directly observe how trajectories starting from a point within the attractor and trajectories starting from its neighborhood diverge, and we numerically estimate the largest Lyapunov exponent.

Let $\{x_0^{(1)}, \ldots , x_0^{(980)}\}$ be the set of 980 test images of the digit~0 as initial values. To remove the transient period before the dynamics settle into the attractor, we map each point for $T=2000$ steps using $G$ and denote the resulting set of points as $X_T = \{x_T^{(1)}, \ldots , x_T^{(980)}\}$. We consider the trajectories starting from these points as reference trajectories and observe the difference between these trajectories and those starting from perturbed points.
For each point $x_T^{(j)}$, the perturbation is applied by selecting the nearest point $x_T^{(k)}$ from the set $X_T$ (excluding itself) and setting 
$y_T^{(j)} = x_T^{(j)} + \varepsilon (x_T^{(k)} - x_T^{(j)})/||x_T^{(k)} - x_T^{(j)}||$, where $\varepsilon=10^{-5}$ is the strength of the perturbation. The intention with this approach is to apply the perturbation in the direction along which the attractor is expanding locally.

We then calculate the difference between these two trajectories as they are transformed by $G$:
\begin{equation}
d_n^{(j)} = ||G^n(y_T^{(j)}) - G^n(x_T^{(j)})||, \label{eq:distance}
\end{equation}
and we compute the sample mean of the logarithm of these values for each step, i.e.,
\begin{equation}
  \bar{d}_n = \frac{1}{980}\sum_{j=1}^{980} \log d_n^{(j)}.
\end{equation}

\begin{figure}[!h]
  \includegraphics[scale=0.7]{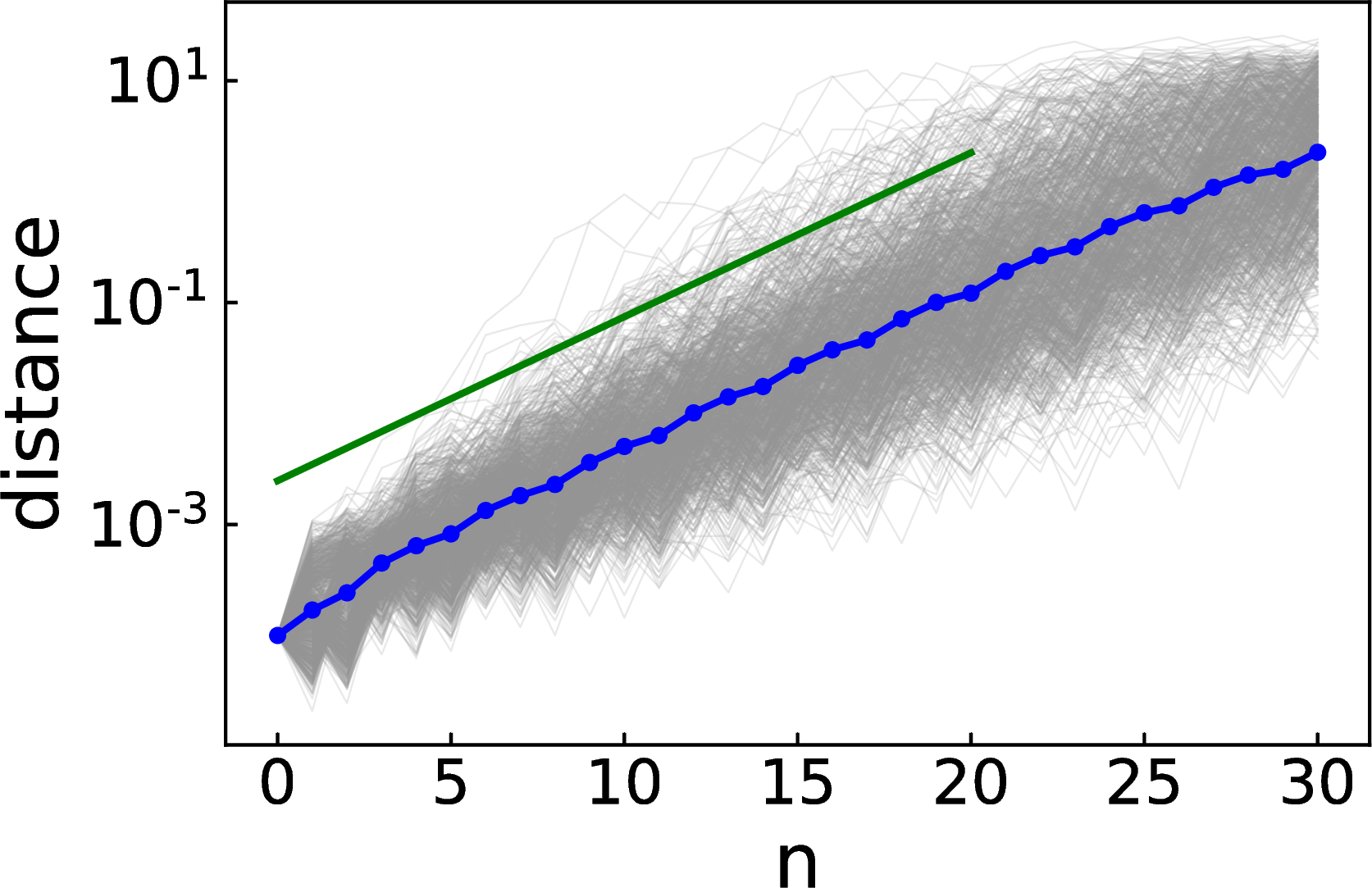}
\caption{{\bf Direct observation of trajectory instability.}
Shown here is the expansion of the differences between the perturbed and reference trajectories and their average. The slope of the green line represents the largest Lyapunov exponent estimated in the previous subsection. The gray lines represent the individual development of the difference between trajectories. The blue line represents the average of the differences.
}
\label{fig:lyapunov_direct}
\end{figure}

Fig~\ref{fig:lyapunov_direct} shows the expansion of the differences between the perturbed and reference trajectories and their average. The slope of the green line represents the largest Lyapunov exponent estimated in the previous subsection. As estimated by linear regression, the actual expansion rate of the errors is 0.352, which is in good agreement with this largest Lyapunov exponent. This consistency between the two approaches indicates that estimating the Lyapunov exponents via the Jacobian matrix provides reliable results.

\subsubsection*{Lyapunov dimension}

In the machine learning community, real-world data such as images are assumed to be distributed on a relatively low-dimensional manifold within the high-dimensional space in which they are embedded. 
The dimension of this manifold is called the intrinsic dimension~\cite{camastra2016intrinsic,pope2021,facco2017}. When generating images using trajectories of a dynamical system, 
if the attractor's dimension matches the intrinsic dimension of the training data, 
then this is considered advantageous for generating a set of images with a diversity similar to that of the original set.  
We characterize the diversity of the generated images by estimating the dimension of the attractor, which can be calculated using the Lyapunov dimension~\cite{alligood2000chaos}.

When the Lyapunov exponents $\lambda_{i}$ $(i=1, \ldots ,N)$ are arranged in descending order, 
and $j$ is an integer satisfying ${\displaystyle \sum_{i = 1}^{j}} \lambda_{i}\geq 0$ and ${\displaystyle \sum_{i = 1}^{j + 1}} \lambda_{i}< 0$,
the Lyapunov dimension $D_L$ is defined as 
\begin{equation}
D_{L}=j+{\displaystyle \frac {\displaystyle {\sum_{i=1}^{j}\lambda_{i}}}{\displaystyle {|\lambda_{j+1}|}}}.
\label{eq:lyapunov_dimension}
\end{equation}
Based on the results in Fig~\ref{fig:lyapunov}, the Lyapunov dimension is estimated to be ca.\ $14.5$. 
According to the literature, the intrinsic dimension of the MNIST dataset is between $10$ and $20$. Although the specific value depends on the method used to calculate it~\cite{pope2021,facco2017}, this range is qualitatively consistent with the result for the Lyapunov dimension. 
These results suggest that images are generated on an attractor by a chaotic dynamical system, which is thought to contribute to the diversity of the images.

\section*{Discussion}
\label{sec:discussion}
In this study, we extended CycleGAN to construct a model that generates images of 
multiple categories by cyclically transforming images among three different categories. 
Using the constructed model, we repeatedly generated images and confirmed that they were 
transformed into images of the following categories. By visualizing the 
distribution of the generated images using a dimensionality reduction technique, 
we verified that the images of the generated data were distributed into three clusters corresponding 
to the same categories as the training dataset.

The process of successive image transformation can be considered as being a dynamical system. 
A single trajectory of the dynamical system induced by our model can generate a diverse 
range of images using chaotic dynamics. Attractors with trajectories that transitioned cyclically among the three 
different categories emerged, producing various images without falling into fixed points or 
periodic solutions. This characteristic is considered effective as a method for generating diverse data.

The quality of the generated images was evaluated using the P/R metric, 
and the precision showed high performance. 
The high precision suggests that the model can accurately capture and reproduce image features. 
However, the recall values were relatively low, 
indicating that the generated images only partially cover the wide distribution of the actual dataset. 
Evaluating these results using precision and recall allowed for a quantitative assessment of 
the outcomes and provided a benchmark for future improvements.

  We conducted a visual investigation of the generated images to address whether the samples within the attractor possess any specific features that render them distinct from the samples outside the attractor. However, it was difficult to characterize the images within the attractor as either typical or atypical samples of their respective categories.

The estimation of the Lyapunov spectrum suggested that the trajectories of the generated images 
exhibit sensitive dependence on initial conditions, a characteristic of chaotic systems. 
This sensitivity was further confirmed by direct observation of trajectories departing from 
neighboring points and diverging from each other. Furthermore, we estimated the dimension of the 
attractor by the Lyapunov dimension, which is considered as the dimension of the data manifold on which the generated images lie. 
The estimated dimension of the attractor was close to the intrinsic dimension of the training dataset. 
This result suggests that the images generated by the model were spread on attractors with a high-dimensional complexity similar to that of the training dataset, and that chaotic dynamics contribute to the diversity of the generated images.

This emergence of the large chaotic attractor may be related to our dynamical system's design, which learned a cyclic path using only the generator $G$, without involving the other generator $F$.
For comparison, consider a dynamical system constructed using the original CycleGAN~\cite{zhu2017unpaired}, defined as $x_{n+1} = F(G(x_n))$. In this case, the composition $F \circ G$ is trained to approximate the identity mapping. Consequently, the trajectory is expected to converge to a small region, such as a fixed point, where no significant changes in the image are expected.
In contrast, we constructed a system that circulates among three categories using only $G$, without being pulled back by $F$. We hypothesize that one of the primary reasons for the emergence of the large chaotic attractor is that this structure is not constrained by the requirement to approach the identity map.
Further comparison of our model's dynamics with those of CycleGAN, and a deeper understanding of the mechanisms underlying the emergence of the chaotic attractor, remain important topics for future research.

Our model can be viewed as an extension of the classical associative memory model that memorizes  sequences of patterns using the Hebbian learning rule~\cite{amari72learning,nara1992chaotic}. 
Classical associative memory can memorize periodic solutions that cycle through multiple  memorized points in the state space of the dynamical system, while the present model cycles among categories instead of points. 
In other words, we have demonstrated that it is possible to construct a model that achieves  hetero-associations among categories. 
Such a model may offer an interesting tool for the interdisciplinary field between machine learning and neuroscience, and future research may investigate how deep learning models perform transformations  among categories and whether properties similar to the classical Hebb's association rule can be found in their model.

To investigate the scope of our method with more challenging real-world examples, we believe it is necessary to extend the approach to handle transformations in latent space, similar to techniques used in StyleGAN~\cite{karras2019style} or latent diffusion models~\cite{rombach2022high}.
To link our method to associative memory processes and achieve more flexible image transformations, several advancements are needed. Potential avenues for future research include:
a) Extending the method to allow variation of specific features within a category (e.g., the thickness or inclination of a character) through external conditional inputs.
b) Developing the capability to store and navigate multiple category cycles.
c) Adding functionality to dynamically change the association target based on conditional inputs.

As evident from the visualized distribution and quantitative results of P/R evaluation,  the generated data did not fully cover the entire distribution of the training data. 
Understanding the balance between the quality and diversity of the generated images and  improving diversity while maintaining quality are challenges for future research. 
To address this, adjusting the parameters of the dropout layer and improving the network structure of the generator and discriminator are considered to be effective approaches.
Furthermore, when calculating the loss function, the discriminator's evaluation of the generated images is currently performed based on only the results of a single mapping from the test images. 
Considering that the recall value decreased with each successive mapping from the test images, it is expected that applying the discriminator's evaluation to the results of multiple transformations of the test images and incorporating this into the loss function could improve the recall value.

\section*{Acknowledgments}
The authors would like to thank Ichiro Tsuda and Shigetoshi Nara for valuable discussions and helpful comments.

% \subsection*{Funding}
% 
% This work was supported by JSPS KAKENHI Grant
% Numbers JP20K11985, JP20H04246, and JP23K11256. The funders had no role in study design, data collection and analysis, decision to publish, or preparation of the manuscript.

% \subsection*{Data Availability}
% Source codes for models, analysis, and visualizations can be found at \url{https://github.com/yymgch/cycle-chaos-gan}

%\nolinenumbers

% Either type in your references using
% \begin{thebibliography}{}
% Text
% \end{thebibliography}
%
% or
%
% \bibliography{refers}
% Compile your BiBTeX database using our plos2015.bst
% style file and paste the contents of your .bbl file
% here. See http://journals.plos.org/plosone/s/latex for 
% \begin{thebibliography}{10}

% \bibitem{bib1}
% Conant GC, Wolfe KH.
% \newblock {{T}urning a hobby into a job: how duplicated genes find new
%   functions}.
% \newblock Nat Rev Genet. 2008 Dec;9(12):938--950.

% \bibitem{bib2}
% Ohno S.
% \newblock Evolution by gene duplication.
% \newblock London: George Alien \& Unwin Ltd. Berlin, Heidelberg and New York:
%   Springer-Verlag.; 1970.

% \bibitem{bib3}
% Magwire MM, Bayer F, Webster CL, Cao C, Jiggins FM.
% \newblock {{S}uccessive increases in the resistance of {D}rosophila to viral
%   infection through a transposon insertion followed by a {D}uplication}.
% \newblock PLoS Genet. 2011 Oct;7(10):e1002337.

% \end{thebibliography}

\section*{Supporting information}

\paragraph*{S1 Fig.}
\label{S1_Fig}
{\bf Model architecture of generator.} Each box represents a layer of the generator and shows the name and type of the layer and the input and output sizes.

{\centering
\includegraphics[scale=0.3]{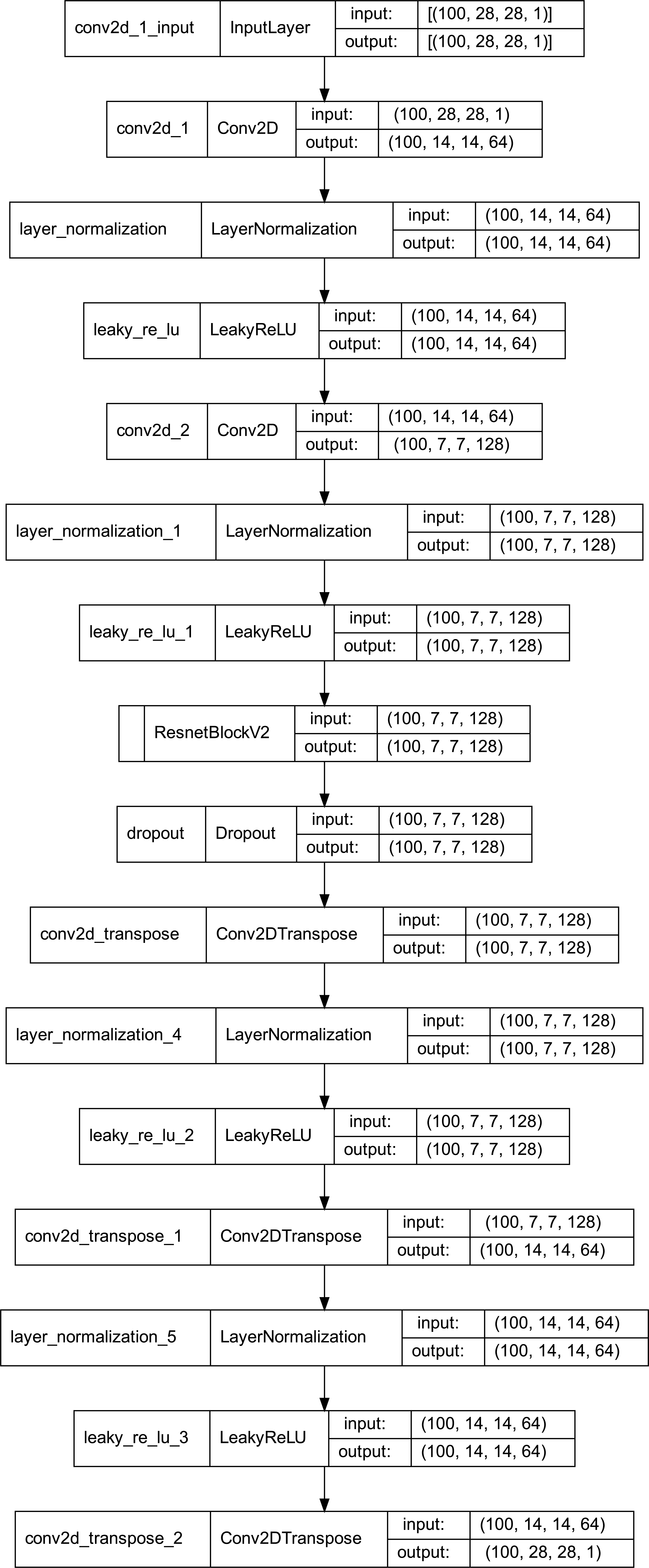}
}

\paragraph*{S2 Fig.}
\label{S2_Fig}
{\bf Model architecture of discriminator.} Each box represents a layer of the discriminator and shows the name and type of the layer and the input and output sizes.

{\centering
  \includegraphics[scale=0.3]{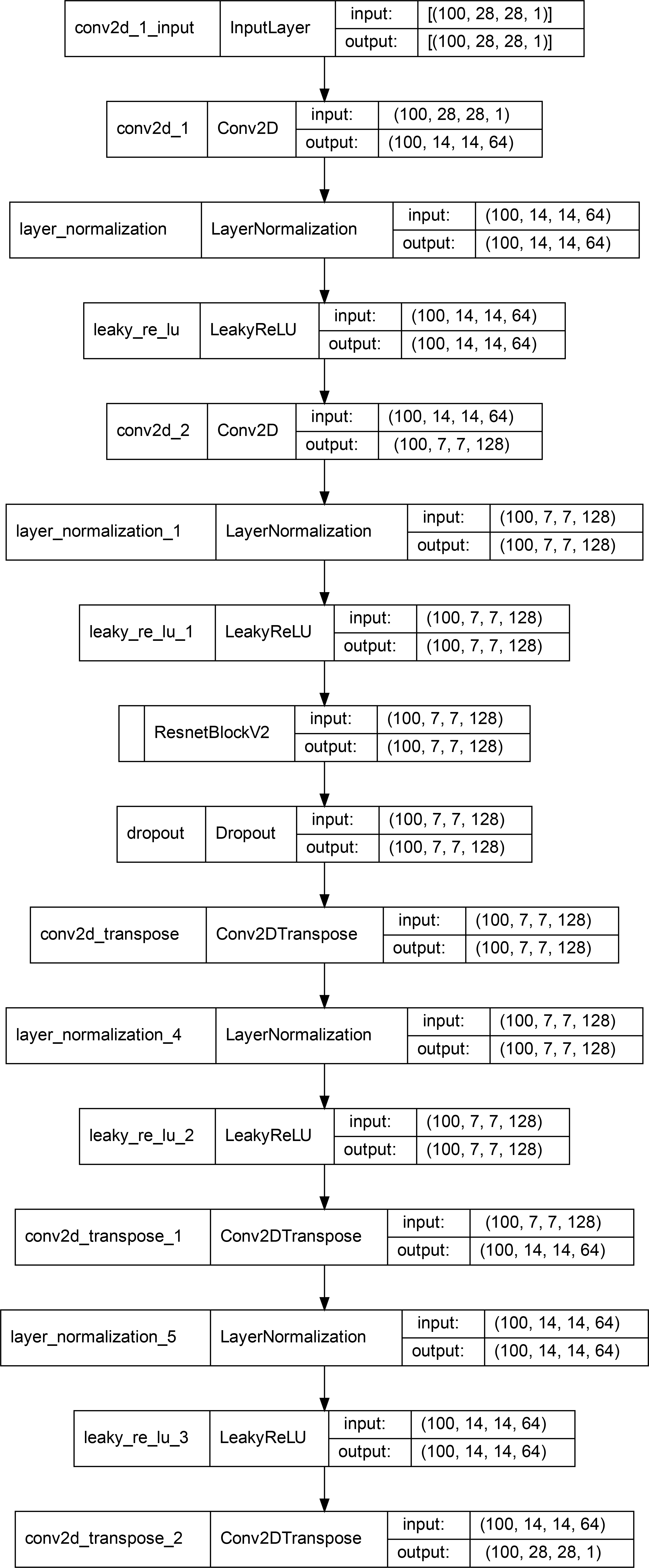}
}

\paragraph*{S3 Fig.}
\label{S3_Fig}
{\bf Trajectories starting from different initial points.}

{\centering
\includegraphics[scale=0.2]{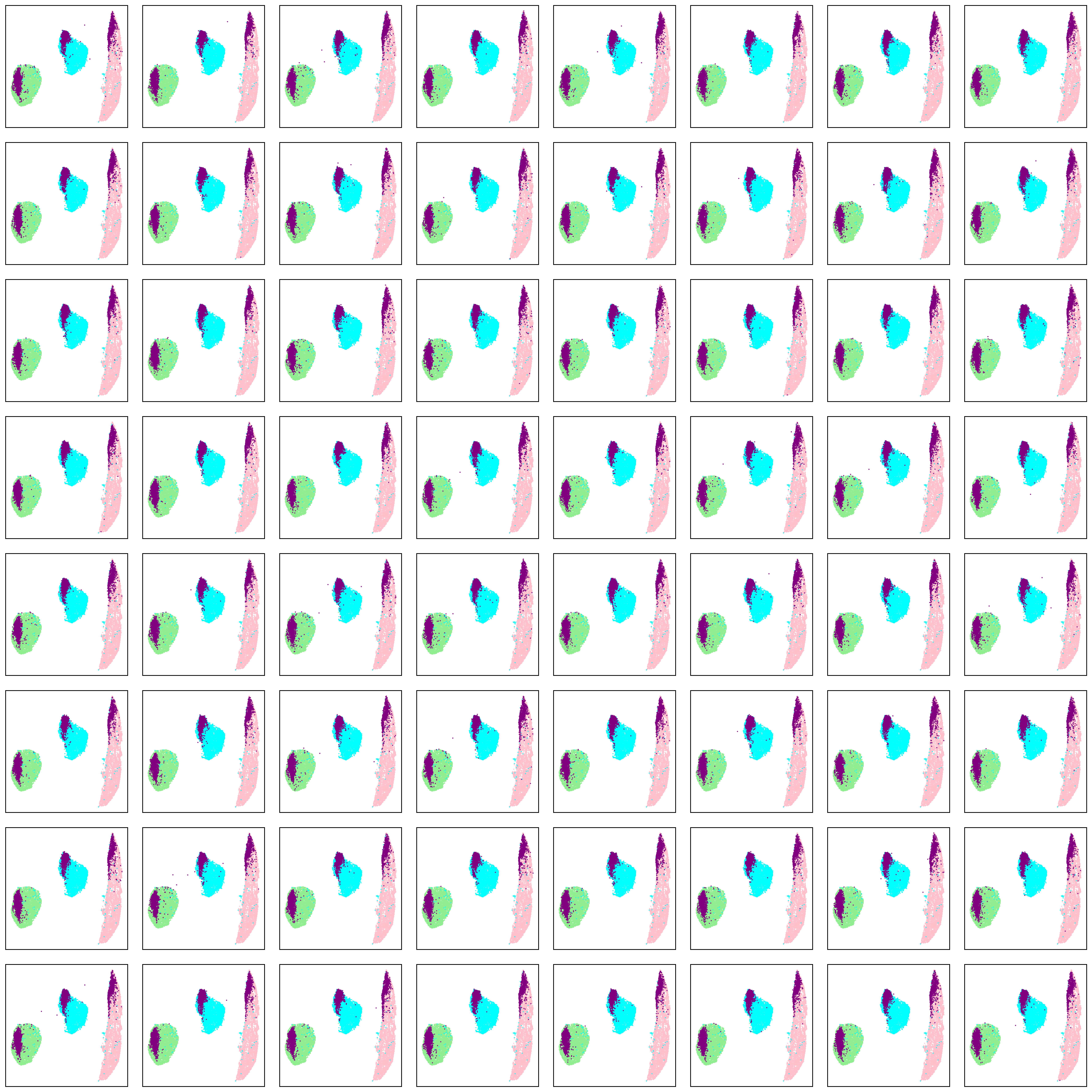}
}

\paragraph*{S4 Fig.}
\label{S4_Fig}
{\bf Precision and recall as functions of time step, calculated from trajectories from perturbed initial images.}
To investigate the dependence of the number of  generated sequences on the evaluation of P/R values, we conducted an analysis using a large number of artificially generated initial images and their trajectories. We perturbed the initial images using the same method as in our maximum Lyapunov exponent analysis (Fig \ref{fig:lyapunov_direct}) and generated $15735$ trajectories ($3147 \times 5$). 
  The strength of the perturbation $\varepsilon$ was set to $0.1$.
  We then evaluated the Precision/Recall metrics using these trajectories and perturbed test images following the same procedure as in Fig \ref{fig:precision_recall_step}.

{\centering
  \includegraphics[scale=0.5]{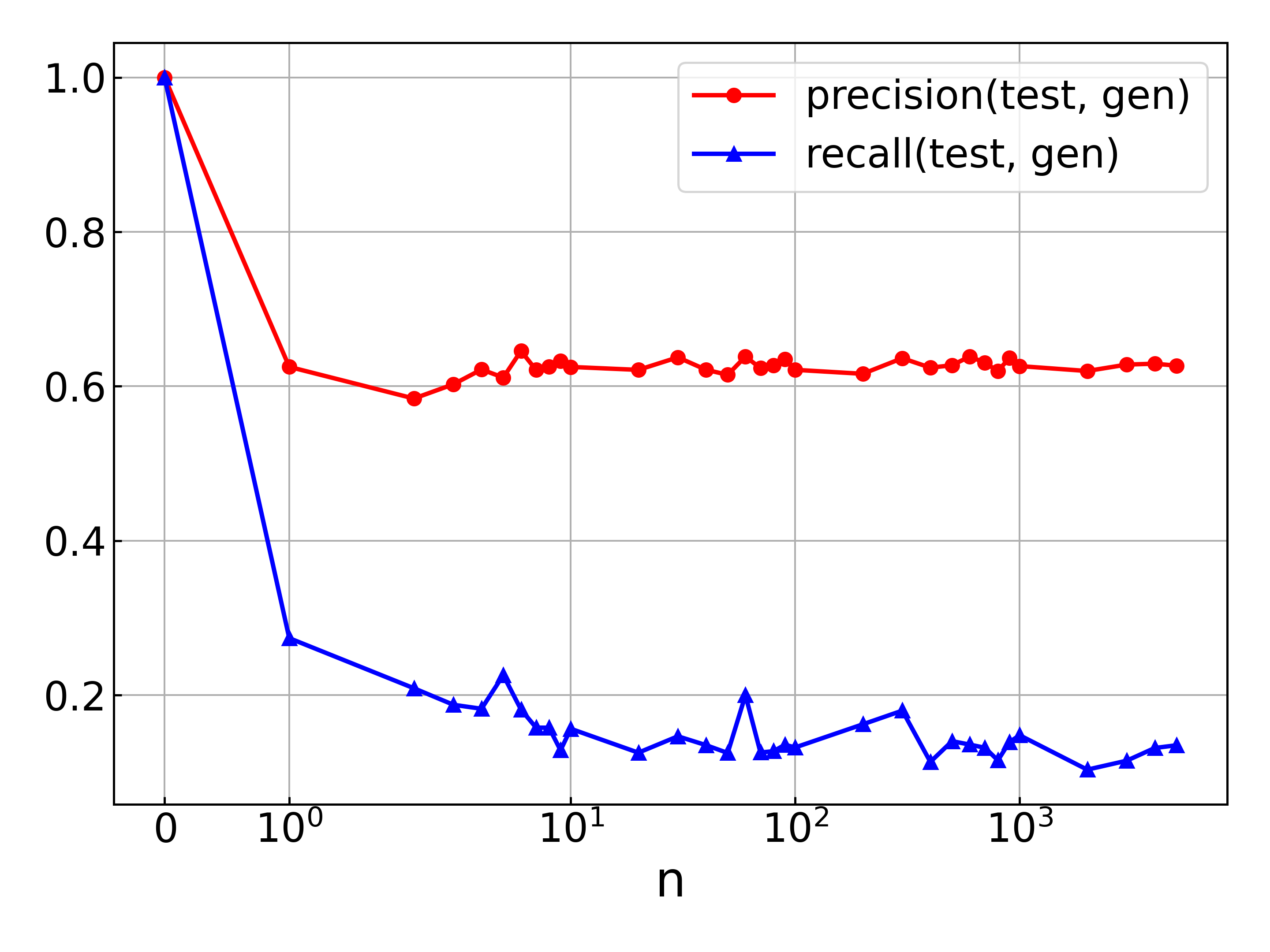}
}

\paragraph*{S5 Fig.}
\label{S5_Fig}
{\bf Histograms of first five Lyapunov exponents calculated for each of 980 trajectories.} 

{\centering
  \includegraphics[scale=0.5]{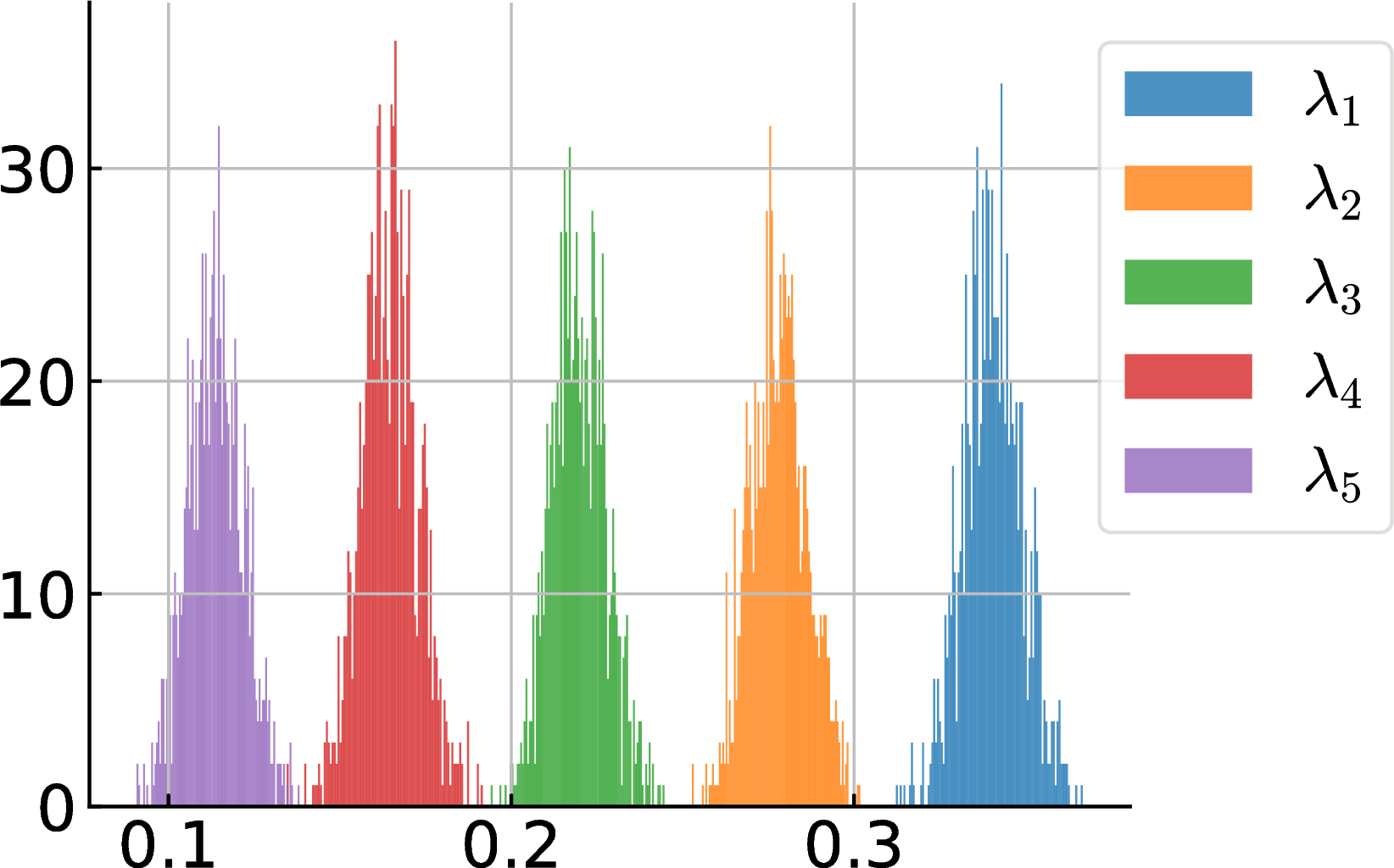}
}

\paragraph*{S6 Fig.}
\label{S6_Fig}
{\bf Convergence of estimated Lyapunov exponents over time.}

{\centering
  \includegraphics[scale=0.5]{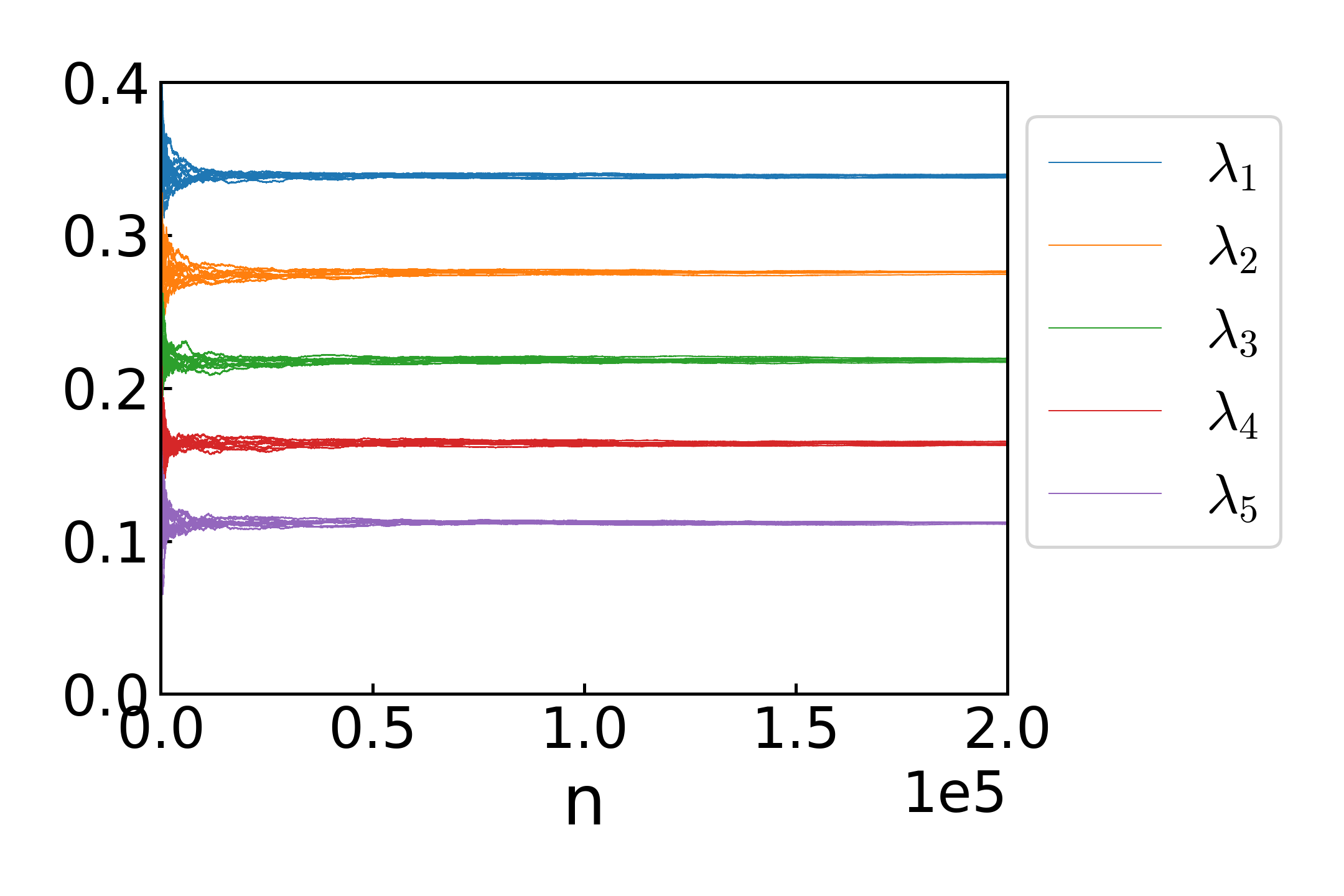}
}

% \bibliography{refers}

\end{document}